\pdfoutput=1 
\documentclass[letterpaper, 10 pt, conference]{ieeeconf}
\IEEEoverridecommandlockouts
\usepackage[figuresleft]{rotating}
\usepackage{amsmath,amssymb,amsfonts}
\usepackage{bm}
\usepackage{hyperref}
\usepackage{caption}
\usepackage{subcaption}
\usepackage{paralist}
\usepackage{graphicx}
\usepackage[ruled,vlined]{algorithm2e}
\usepackage{titlesec}
\usepackage{booktabs}
\usepackage{xcolor}
\usepackage{cases}
\usepackage{fontawesome}
\newcommand{\etal}{\textit{et al.}}

\hypersetup{colorlinks=true,urlcolor=blue,citecolor=red}

\title{\LARGE \bf
Grasping in the Dark: Zero-Shot Object Grasping\\Using Tactile Feedback}

\author{Kanishka Ganguly$^1$, Behzad Sadrfaridpour$^2$, Pavan Mantripragada$^1$, Nitin J. Sanket$^1$,\\ Cornelia Ferm{\"u}ller$^1$, Yiannis Aloimonos$^1$

\thanks{$^1$Authors are associated with the Perception and Robotics Group, University of Maryland, College Park, MD, 20740. Email: {\small{\texttt{\{kganguly, mppavan, nitinsan, fermulcm, jyaloimo\}@umd.edu}}}}%
\thanks{$^2$Author was associated with the Perception and Robotics Group when he made most of his contributions. Email: {\small\texttt{behzad@faridpour.com}}}%
}

\begin{document}

\maketitle
\thispagestyle{empty}
\pagestyle{empty}

\begin{abstract}
Grasping and manipulating a wide variety of objects is a fundamental skill that would determine the success and wide spread adaptation of robots in homes. Several end-effector designs for robust manipulation have been proposed but they mostly work when provided with prior information about the objects or equipped with external sensors for estimating object shape or size. Such approaches are limited to many-shot or unknown objects and are prone to estimation errors from external estimation systems.

We propose an approach to grasp and manipulate previously unseen or zero-shot objects: the objects without any prior of their shape, size, material and weight properties, using only feedback from tactile sensors which is contrary to the state-of-the-art. Such an approach provides robust manipulation of objects either when the object model is not known or when it is estimated incorrectly from an external system. Our approach is inspired by the ideology of how animals or humans manipulate objects, i.e., by using feedback from their skin. Our grasping and manipulation revolves around the simple notion that objects slip if not grasped stably. This slippage can be detected and counteracted for a robust grasp that is agnostic to the type, shape, size, material and weight of the object.
At the crux of our approach is a novel tactile feedback based controller that detects and compensates for slip during grasp.
We successfully evaluate and demonstrate our proposed approach on many real world experiments using the Shadow Dexterous Hand equipped with BioTac SP tactile sensors for different object shapes, sizes, weights and materials. We obtain an overall success rate of 73.5\%.



\end{abstract}
\section{Introduction}
\label{sec:intro}

Robotic agents, and their respective research fields, have generally proven useful in structured environments, crafted specifically for them to operate. We as robotics researchers envision in the near future, robots performing various tasks in our homes. For such robots to be successful when deployed ``in the wild'', they have to grasp and manipulate objects of various shapes, sizes, materials and weight which may or may not be present in their database.


Having the ability to grasp robustly and repeatedly is the primary way by which robots can affect their environment, and is the first step to performing more complicated and involved tasks. Robust grasping involves reliable perception of the object form (inference of pose, type of object and other properties) before grasping and a robust and continuous feedback loop to ensure that object does not slip (or fall) during grasping and manipulation. 

In this work, we focus on the latter since it is required for grasping previously unseen or zero-shot objects, i.e., objects of unknown size, shape, material and weight, utilizing tactile feedback. Our method is inspired by the amazing grasping abilities of animals and humans \cite{Camponogara2019, Breveglieri2016} for novel objects, and how they are able to ``grasp in the dark'' (with their eyes closed). To showcase that our method does not rely on an external perception input but rather only relies on tactile proprioception, we also demonstrate our method working on transparent objects that cannot be sensed robustly using traditional perception hardware, such as cameras or LIDARs.
We formally define our problem statement and summarize a list of our contributions next.

\subsection{Problem Formulation and Contributions}
A gripper is equipped with tactile sensors and an object of unknown shape, size, material and weight is placed in front of the gripper. The problem we address is as follows: \textit{Can we grasp and manipulate an unseen and unknown object (zero-shot) using only tactile sensing?}\\
We postulate that a robust grasp is achieved when enough force is applied to an object such that it is \textit{just sufficient} to counteract gravity, thus suspending the object in a state of static friction experienced between the object and the finger. Our framework allows for robust grasping of previously unseen objects or zero-shot of varied shapes, sizes, and weights under the absence of visual input, due to our reliance solely on tactile feedback. Humans are capable of this feat from an early age~\cite{Clifton1993}, and it is an important ability to have in scenarios with the absence of visual perception due to occlusions or the object not being present in the robot's knowledge-base. A summary of our contributions are:
\begin{itemize}
    \item We propose a tactile-only grasping framework for unseen or zero-shot objects.
    \item Extensive real-world experiments showing the efficacy of the proposed approach on a variety of common day-to-day objects of various shapes, sizes, textures and rigidity. We also include challenging objects such as a soft-toy and a transparent cup demonstrating that our approach is robust.
\end{itemize}

\section{Related Work}
\label{sec:prior_work}


Most of the research in tactile-related grasping can be broadly divided into three categories, those that rely on purely tactile input, those that use vision-based approaches, and those that perform end-to-end learning of grasping in simulation and attempt to transfer them to a real robot. We discuss some of the recent works done in all three categories, and their respective pros and cons.

\subsubsection{Tactile Grasping}
In 2015, \cite{schaal2015force} presented their force estimation and slip detection for grip control using the BioTac sensors where they try to classify ``slip events'' by looking at force estimation for the fingers. They  proposed a grip controller, which helps adapt the grasp if slip is detected. They only used the pressure sensor data and only considered 2 or 3 fingers, and compared it with an IMU placed on the object itself.\\
In 2016, \textit{BiGS: Biotac Grasp Stability} dataset~\cite{chebotar2016bigs} was released, which equipped a Barrett three-fingered hand with BioTac sensors and measured grasp stability on a set of objects, classified into cylindrical, box-like and ball-like geometries. In 2018, \cite{zapata2018nonmatrix} presented their work on non-matrix tactile sensors, such as the BioTac, and how to exploit the local connectivity to predict grasp stability. They introduced the concept of ``tactile images'' and used only single readings of the sensor to achieve high rate of detection compared to multiple sequential readings.
In 2019, TactileGCN~\cite{garcia2019tactilegcn}  was presented. The authors used a graph CNN to predict grasp stability and they used the BioTac sensor data to construct the graph, but used only three fingers to grasp. They employed the concept of ``tactile images'' to convert grasp stability into an image classification problem. Their approach only deals with static grasps and does not consider the dynamic interaction between objects and the fingers. 
Another work \cite{pestell2019sense} tackled this problem by extracting features from high-dimensional tactile images and infer relevant information to improve grasp quality. But their approach is restricted to flat, dome- and edge-like shapes. 
The work by \cite{peters2019tactilelibrary} used FingerVision \cite{FingerVision} sensor mounted on a parallel gripper to generate a set of tactile manipulation skills, such as stirring, in-hand rotation, and opening objects with specified force. However, FingerVision is only appropriate for demonstraing proof of concept since it has a large form factor and is not robust. 
In 2020, a new tactile sensor ``DIGIT'' is presented in \cite{lambeta2020digit} that learns to manipulate small objects with a multi-fingered hand from raw, high-resolution tactile readings.  In \cite{veiga2020tactilegripstabilization}, the authors use the BioTac sensor as a way to stabilize objects during grasp using a grip force controller. The underlying assumption is that the shape of the object is known a-priori and repeatability with different shapes and sizes remains an ongoing challenge.

\subsubsection{Visual Input  (with Tactile Input) Grasping}
In \cite{nakamura2017complexities}, the authors demonstrate a data set of slow-motion actions (picking and placing) organized as manipulation taxonomies. In \cite{calandra2018more}, an end-to-end action-conditioned grasping model is trained in a self-supervised manner that learns re-grasping from raw visuo-tactile data, where the robot receives tactile input intermittently. The work in \cite{handa2020dexpilot} leverages the innovation in machine vision, optimization and motion generation to develop a low-cost glove-free teleoperation solution to grasping and manipulation. 

\subsubsection{Simulation Based Grasping}
Perhaps the most popular and famous papers in this category are \cite{openai2018dexterity, openai2019rubiks} from OpenAI, in which the authors demonstrate a massively parallel learning environment for the Shadow Dexterous Hand, and learn in-hand manipulation of a cube, and the solving of a Rubik's cube entirely in simulation after which they are able to transfer said learning onto a physical robot. While impressive, their transfer learning approach requires near-perfect information about the joint angles of the Hand, as well as visual feedback regarding the position of the object.
Levine, \etal~\cite{levine2016learning} performs large-scale data collection and training on 14 manipulators for learning hand-eye coordination, directly going from pixel-space to task-space.
In \cite{zhu2019dexterous}, a model-free deep reinforcement learning which can be scaled up to learn a variety of manipulation behaviors in the real world has been proposed, using general purpose neural networks. A State-Only Imitation Learning (SOIL) is developed in \cite{radosavovic2020state}, by training an inverse dynamics model to predict action between consecutive states. The research problems attempted using perception and learning has seen limited progress due to the fact that vision does not provide any information regarding contact forces, regularly fails to reconstruct the scene due to occlusion or that the material properties of the object and the process of learning is time consuming, requires large amounts of data, and sometimes does not transfer to a real robot \cite{liu2020skill}.

\subsection{Organization of the paper}
We provide an overview of our approach in Sec. \ref{sec:approach} followed by detailed description of our hardware setup and software pipeline in Secs. \ref{sec:hwsetup} and \ref{sec:swdesign}. We then present our experiments in Sec. \ref{sec:experiments} along with an analysis of the results in Sec. \ref{sec:analysis}. Finally, we conclude our work in  Sec.~\ref{sec:conclusions} with parting thoughts for future work.

\section{Overview}
\label{sec:approach}
\begin{figure*}[t!]
\centering
    \includegraphics[width=1.0\linewidth]{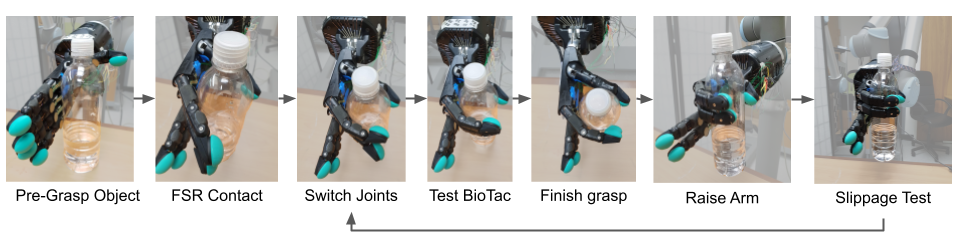}
    \caption{Grasp pipeline demonstration.}
    \label{fig:pipeline}
    \vspace{-5px}
\end{figure*}

\begin{figure}[t!]
    \centering
    \includegraphics[width=1.0\linewidth]{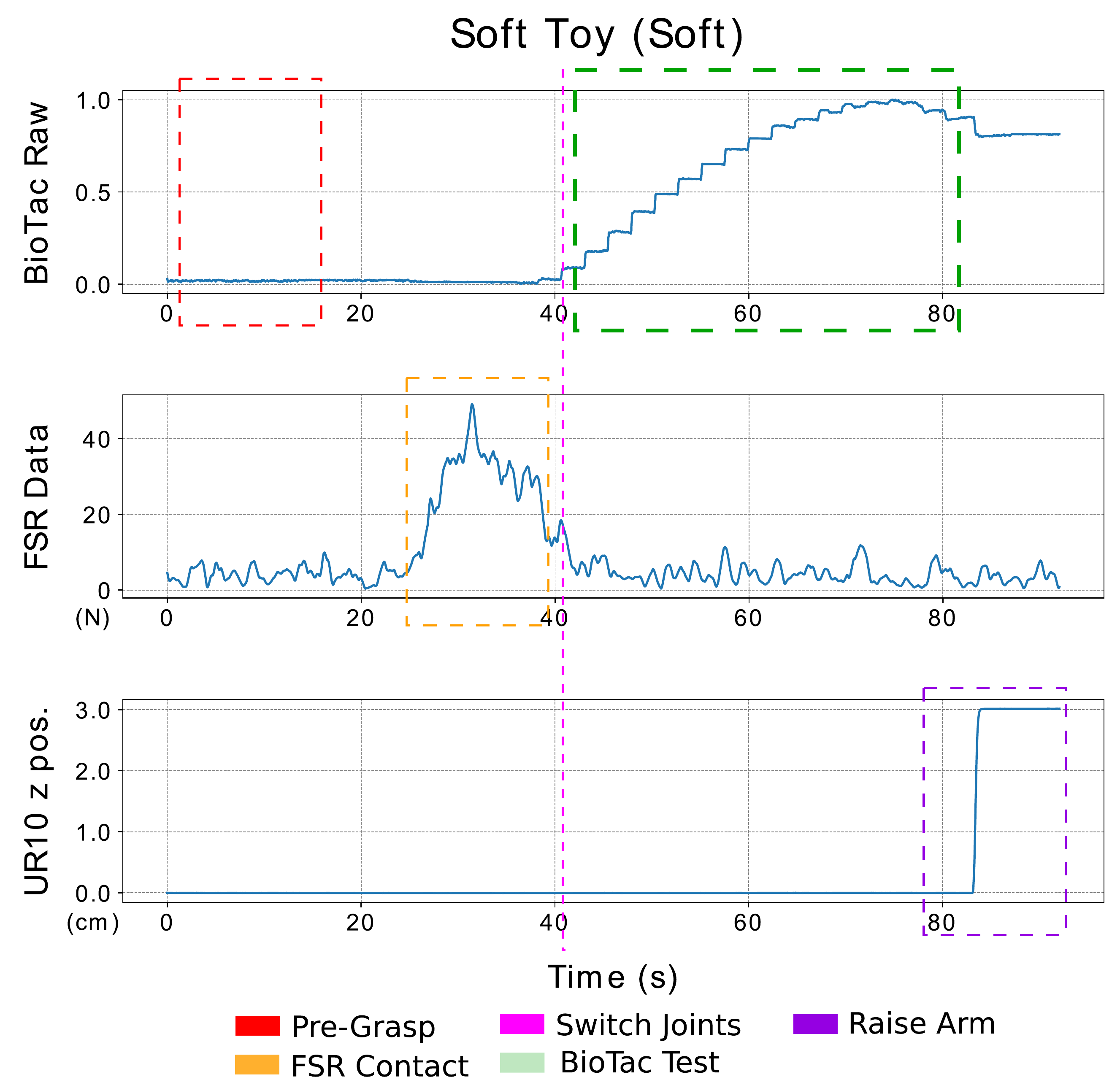}
    \caption{Variation of BioTac and FSR data as the soft toy is grasped and lifted from the table. \textit{All the graphs in this paper are best viewed in color on a computer screen at a zoom of 200\%}}
    \label{fig:pipeline_plot}
    \vspace{-0.3cm}
\end{figure}

Our approach to solving the problem of grasping zero-shot objects is to define different actions for the robot and execute them accordingly. Figs.~\ref{fig:pipeline} and \ref{fig:pipeline_plot} shows the implementation and plots of these actions. An overview for controlling the robot to execute each of these actions are described in the rest of this section. Our proposed execution approach is implemented on a combination of the Shadow Dexterous Hand (we will call this ShadowHand, for brevity) equipped with BioTac SP tactile sensors (we will call this BioTac, for brevity) attached to a UR-10 robotic manipulator. An overview of our approach is as follows:
\begin{itemize}
    \item \textit{FSR Contact}: Control each finger such that its proximal phalanges (the phalanges nearest to the palm) reaches the object.
    \item \textit{Switch Joints}: Control each finger's configuration such that its distal phalanges (the fingertip) reaches the object.
    \item \textit{Raise Arm}: Move the robotic arm configuration upwards while controlling the robotic hand's configuration to prevent object from slipping.
\end{itemize}

Before understanding our grasping framework, i.e. slip detection followed by a control policy for slip compensation, it is important  to  understand  the  basic  structure  of the ShadowHand. This will help the reader gain an intuition about the formulation of our control policy. The hardware setup is explained in the Sec. \ref{sec:hwsetup} followed by our software pipeline in Sec. \ref{sec:swdesign}.

\section{Hardware Setup}
\label{sec:hwsetup}

\subsection{Kinematic Structure of the Shadow Dexterous Hand}

\begin{figure}[ht!]
    \centering
    \includegraphics[width=1.0\linewidth]{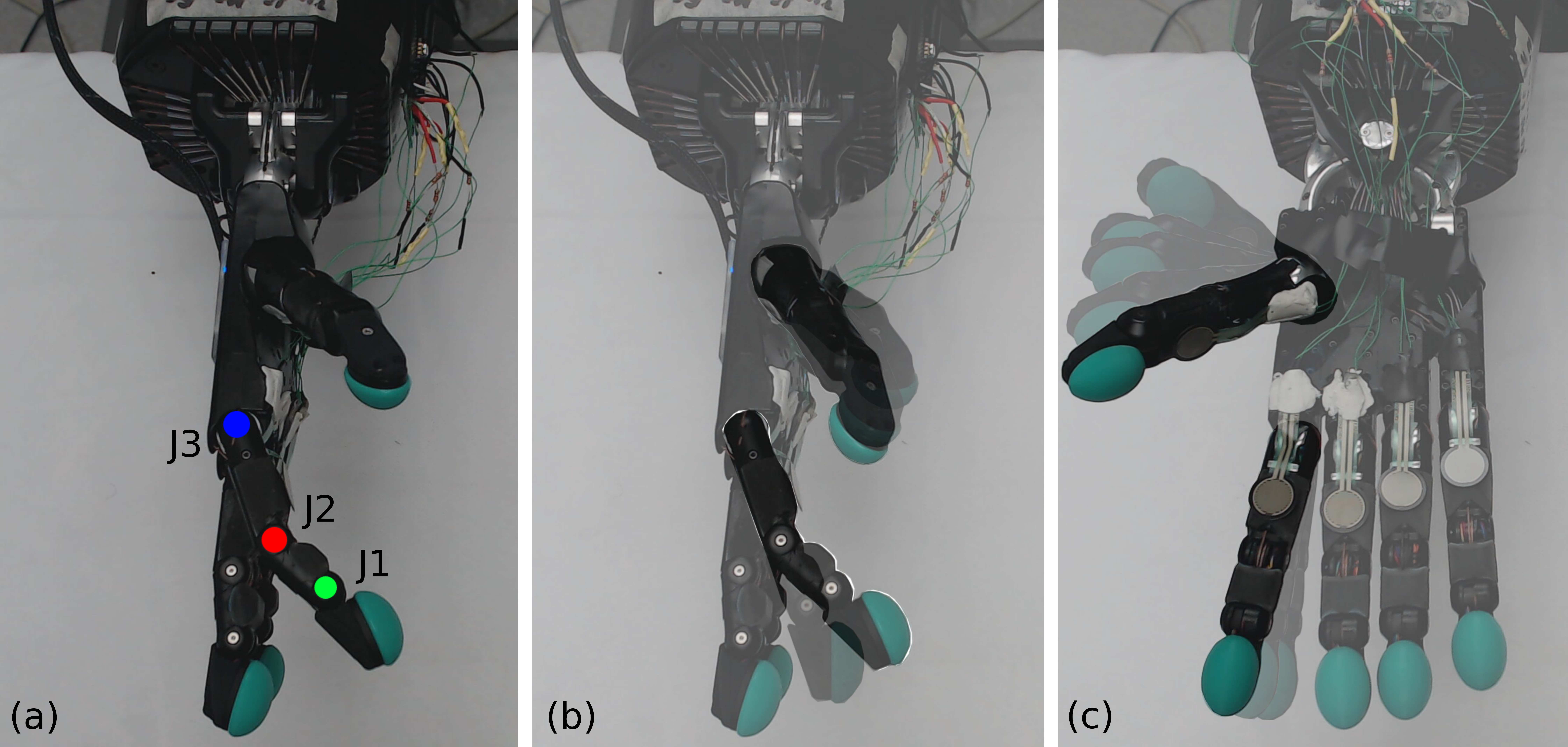}
    \caption{ShadowHand (a) joint nomenclature and (b, c) finger and thumb joints with their limit positions.}
    \label{fig:finger_diagram} \vspace{-0.3cm}
\end{figure} 

The ShadowHand has four fingers and an opposable thumb. Each of the fingers have four joints while the thumb has five joints. A representation of motion is shown in  Fig.~\ref{fig:finger_diagram}.

Fig.~\ref{fig:finger_diagram}\textcolor{red}{a} demonstrates one finger and the joints specification it follows. Each finger has three links, also called phalanges, with one joint in between. From the top of the finger to the base, these are called the distal, middle and proximal phalanges respectively.\\
The fingers can be controlled by sending joint position values. The joint $J_3$ has two controllable ranges of motion, along the \textit{sagittal} and \textit{transverse} axes. This joint also has a minimum and maximum range of $0^\circ$ to $90^\circ$ respectively. The joints $J_1$ and $J_2$ are different in that, similar to the human hand, they are coupled internally at a kinematic level and \textit{do not move independently}. They individually have a range of motion between $0^\circ$ to $90^\circ$, but are underactuated. This means that the angle of the middle joint, i.e. $J_2$ is always greater than or equal to the angle of the distal joint, i.e. $J_1$ which allows the middle phalanx to bend while the distal phalanx remains straight. 

\begin{figure}[ht!]
    \centering
    \includegraphics[width=0.7\linewidth]{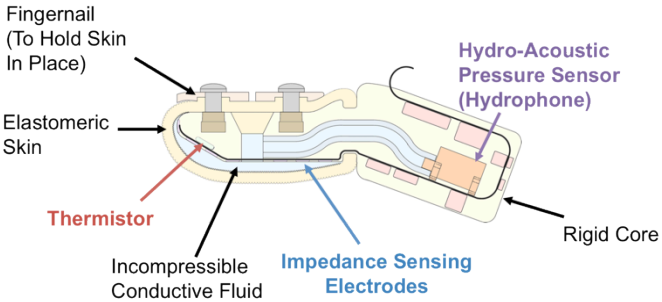}
    \caption{Cross-section of the BioTac sensor.}
    \label{fig:biotac_internal}
\end{figure}

Currently available commercial tactile sensors lie on a spectrum spanning from accuracy on one end to form factor on the other. These sensors can either have high accuracy while sacrificing anthropomorphic form factor or can be designed similar to human fingers, while having a relatively poor accuracy at tactile sensing. The choice of the sensor depends on the task at hand, which in our case is to grasp zero-shot objects. To this end, we select the ShadowHand equipped with the BioTac sensors in an effort to be biomimetic. 

Using a combination of impedance sensing electrodes, hydro-acoustic pressure sensors and thermistors, the BioTac sensor is capable of sensing three of the most important sensory inputs that one needs for grasping, namely deformation and motion of stimuli across the skin, the pressure being applied on the finger and temperature flux across the surface. The internal cross-section of the BioTac is shown in Fig.~\ref{fig:biotac_internal}.
\begin{figure}[ht!]
    \centering
    \subfloat[FSRs\label{fig:fsr_hand}]{\includegraphics[width=.13\textwidth]{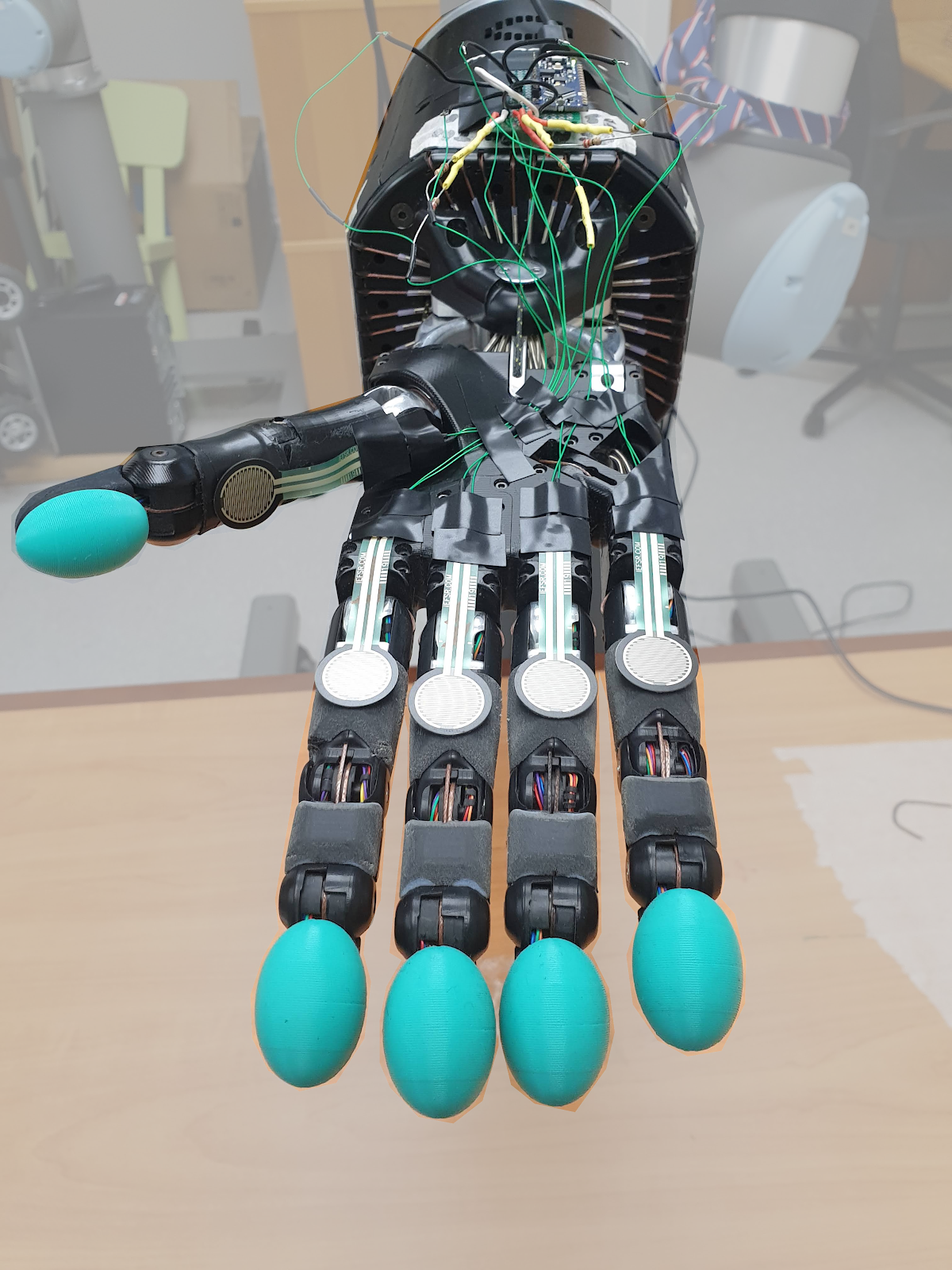}}\quad
    \subfloat[Connections\label{fig:arduino}]{\includegraphics[width=.13\textwidth]{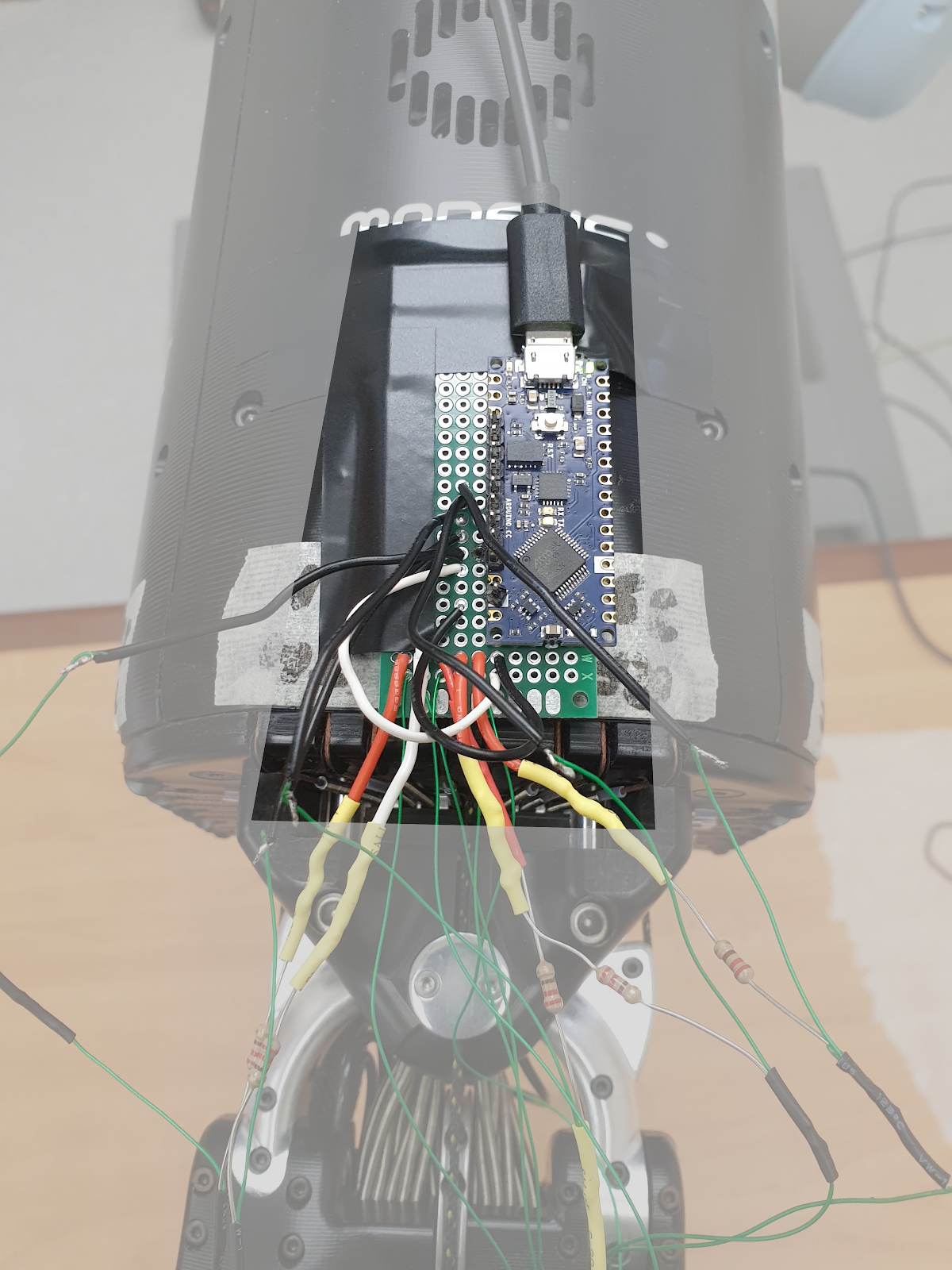}}\quad
    \subfloat[Contact Regions\label{fig:palm_contacts}]{\includegraphics[width=.12\textwidth]{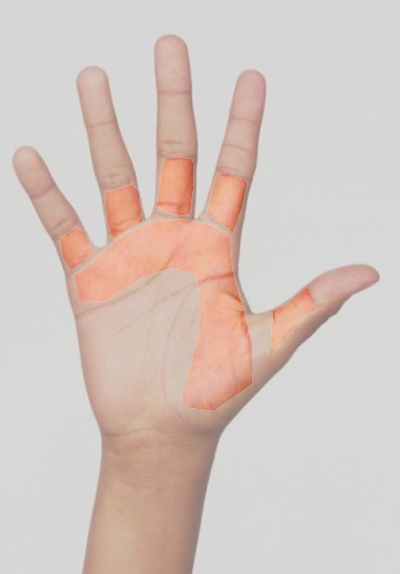}}
    \caption{(a) FSRs, (b) FSR connection to Arduino Nano, and (c) Regions of contact when grasping.}
\end{figure} 

The human hand can grasp objects of various shapes, sizes and masses without having seen them previously. This ability to grasp previously unseen objects in the absence of visual cues is possible only due to the presence of tactile sensing over a large surface area, through the skin. In Fig.~\ref{fig:palm_contacts}, the highlighted parts show the primary regions of contact when grasping is performed. These regions make first contact with the object being grasped and apply the most amount of force, due to the large surface area. To mimic similar tactile characteristics on the ShadowHand, we equip it with additional sensors at the base of each finger and the thumb.\\
We utilize Force Sensitive Resistors (FSRs) for this purpose, which are flexible pads that change resistance when pressure is applied to the sensitive area (See Fig.~\ref{fig:fsr_hand}). These FSRs work on the principle of a voltage divider circuit, Fig.~\ref{fig:arduino}, and have a voltage drop inversely proportional to the resistance of the FSR.

We calibrate our sensors using a ground-truth force measurement unit, for 0N to 50N of force, using simple regression. We map a series of readings from the FSR to the corresponding force values in Newtons on the force measurement unit, and fit a regression line to these points. This is sufficient to measure contact forces between the fingers and an object during grasping.

\section{Software Pipeline}
\label{sec:swdesign}

\subsection{Grasp Controller} \vspace{-0.1cm}

Our pipeline, defined in Algos.~\ref{alg:grasp},~\ref{alg:lift}, starts at the pre-grasp pose where the fingers are fully extended and the thumb is bent at the base to a $70^\circ$ angle, which is optimal for grasping most objects due to having the maximum volume coverage by the trajectories of the finger tips. We explain the distinct parts of our pipeline in the following sections.

\subsubsection{Initialization}
The controller begins by performing a tare operation using 50 readings of each BioTac sensor and computing their respective means. Successive readings are then min-max normalized, within an adjustable threshold of $\pm200$ of this mean, to ensure that each sensor's biases are taken into account, as well as to provide a standardized input to the control loop.

\subsubsection{Hand-Object Contact}
Once the initialization process is complete and baseline readings have been established, the hand controller begins actuating the $J_3$ joints of all the fingers and $J_4$ of the thumb. This is done by sending the appropriate joint control commands. The current joint values are obtained from the ShadowHand, checked against the maximum joint limits of each finger ($90^\circ$ for $J_3$), and increased by a small angle $\delta \theta$. The $J_3$ and $J_4$ joints of the fingers and thumb respectively are moved until it registers a contact with the object, as measured by the FSR readings. This establishes an initial reference point for the hand to begin refining the grasp, and the controller switches to the control policy for the coupled joints.

\subsubsection{Preliminary Grasp}
At this stage, since the base of each finger and thumb have made initial contact with the object, the control policy switches to the coupled joints so that the fingers can begin to ``wrap around'' the object. We now activate the coupled $J_1$ and $J_2$ joints (controlled using a virtual $J_0$ joint defined in Eq. \ref{eq:j0}) of the fingers and the $J_1$ joint on the thumb. In this stage, the $\delta \theta$ is inversely proportional to the the normalized sensor data from the BioTac sensor.

\begin{equation}
J_0 = \begin{cases}
J_2 \quad\text{if}\quad J_2\in [0,90]^\circ
&\\
J_1 \quad\text{if}\quad J_2>90^\circ\\
\end{cases}
\label{eq:j0}
\end{equation}



Intuitively, this means that when there is little or no contact between the fingers and the object, the controller sends out larger joint angle targets, causing the fingers to move larger distances. Once contact is made, the controller moves the fingers at progressively smaller increments, thus allowing for a more stable and refined grasp.

We use a Bezi\'er curve to generate an easing function that maps our normalized BioTac sensor data to a normalized angle (in radians), between the joint limits of the respective joint. This mapping is then converted into a usable control output $\delta \theta \in [\theta_{\textrm{min}}, \theta_{\textrm{max}}]$.

The Bezi\'er curve is generated by the parametric formula 

\begin{equation}
\label{eq:bezier}
\theta_{\beta} = S_{\textrm{BioTac}}^2 \times (\kappa_1 - (\kappa_2 \times S_{\textrm{BioTac}}))
\end{equation}

where $\kappa_1$ and $\kappa_2$ are the Bezi\'er control points, $S_{\textrm{BioTac}}$ is the instantaneous normalized BioTac reading and $\theta_{\beta}$ is the mapped Bezi\'er curve output. We then compute control output $\delta \theta$ as follows

\begin{equation}
\label{eq:control_output}
\delta \theta = B_1 + \frac{(\theta_{\beta} - A_1)\times (B_2-B_1)}{A_2-A_1}
\end{equation}

where $[A_2, A_1] \in [0,1]$ and $[B_2, B_1] \in [\theta_{\textrm{min}},\theta_{\textrm{max}}]$.\\
We set a termination threshold $\tau_{\textrm{termination}} = 0.1$ on the BioTac sensor values such that the Hand controller stops executing as soon as a minimal level of contact is detected. Once all the fingers and the thumb have reached the preliminary grasp state, we exit the control loop.

\begin{algorithm}
    \caption{Implementation of the initial grasp controller}
    \SetAlgoLined\DontPrintSemicolon
    \SetKwFunction{FnReset}{Reset}
    \SetKwProg{Fn}{Procedure}{:}{}
    \Fn{\FnReset}{
        \Repeat{pre-grasp reached}{
            Move Hand to pre-grasp pose\;
        }
    }
    Baseline = $\mathrm{mean}_{50}$ of BioTac\;
    \Repeat{Until FSR contact}{
        $J_3 = J_3 + \delta \theta$\;
        Actuate $J_3$
    }
    Switch to Coupled Joints Controller\;
    Compute control output based on Eqs.~\ref{eq:bezier},~\ref{eq:control_output}\;
    \Repeat(\tcc*[f]{\footnotesize{Until fingertip touches object}}){Until $P_{\textrm{dc}}^{t} - P_{\textrm{dc}}^{t+1} \geq \tau_{\textrm{termination}}$}{
        \While{$J_1 + J_2 \leq 180^\circ$}{
        $J_2 = J_2 + \delta \theta$\;
            \eIf(\tcc*[f]{\footnotesize{Joint limit}}){$J_2^{t} - J_2^{t+1} < 0.1 $}{
                 $J_1 = J_1 + \delta \theta$\;
                 Actuate $J_1$\;
            }{
                Actuate $J_2$\;
            }
        }
    }
    \label{alg:grasp}
\end{algorithm}
\vspace{\baselineskip}
\setlength{\textfloatsep}{0pt}
\begin{algorithm}
    \SetAlgoLined\DontPrintSemicolon
    \SetKwFunction{FnSlip}{SlippageDetection}
    \SetKwProg{Fn}{Procedure}{:}{}
    \Fn{\FnSlip}{
        \While{Hand has not been raised}{
            \Repeat{slip detected}{
                Actuate $J_1$\;
                Move UR-10 up\;
            }
        }
    }
    \caption{Implementation of picking with slip detection}
    \label{alg:lift}
\end{algorithm}

\begin{figure}[ht!]
    \centering
  {\includegraphics[width=1.0\columnwidth]{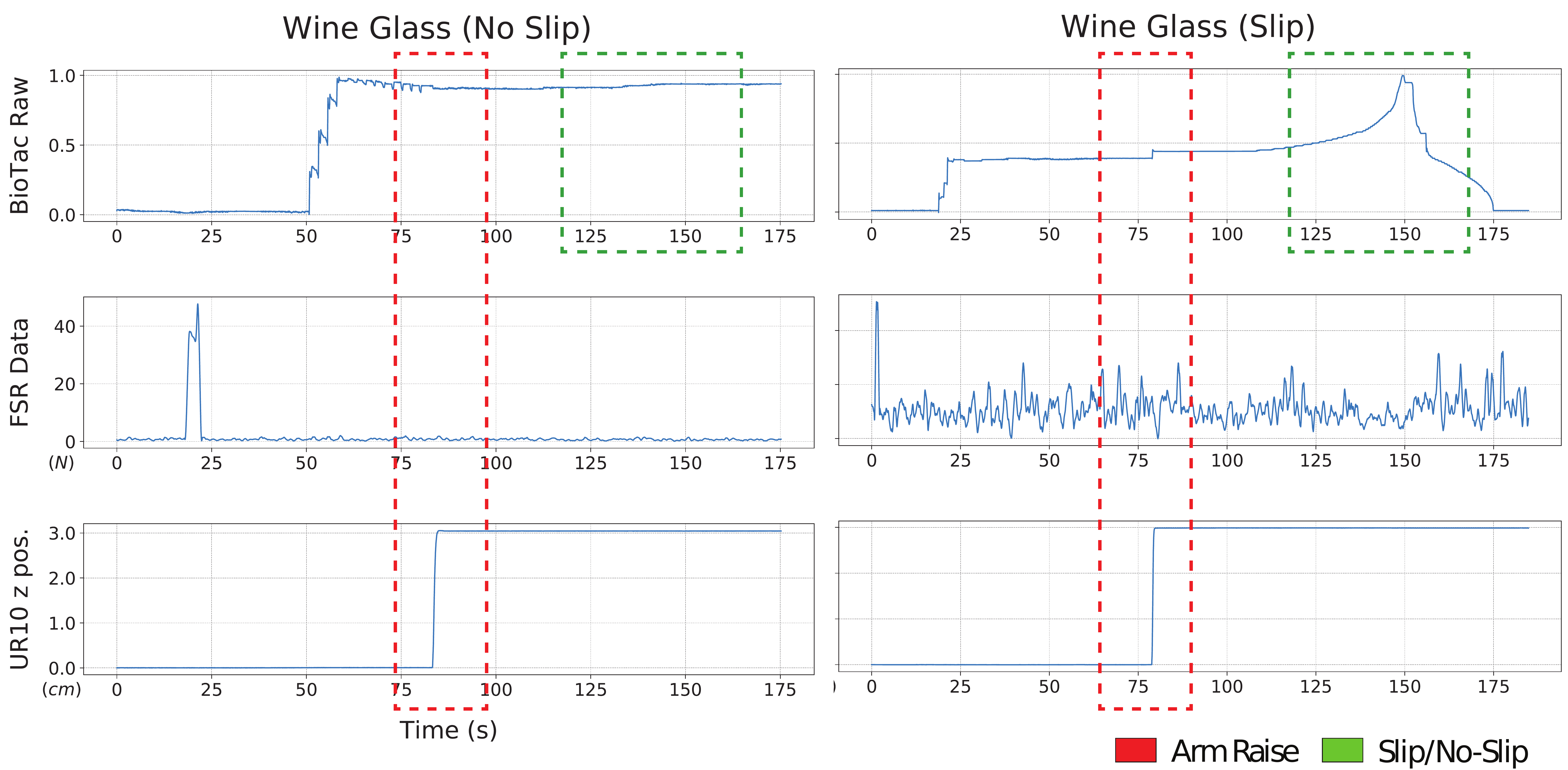}}
    \caption{Plot of BioTac and FSR sensor data for without and with slip on the wine glass experiment.}
    \label{fig:grasp_slip_combo}
\end{figure}

\subsection{Slippage Detection}

\begin{figure}[ht!]
    \centering
    \includegraphics[width=0.5\textwidth]{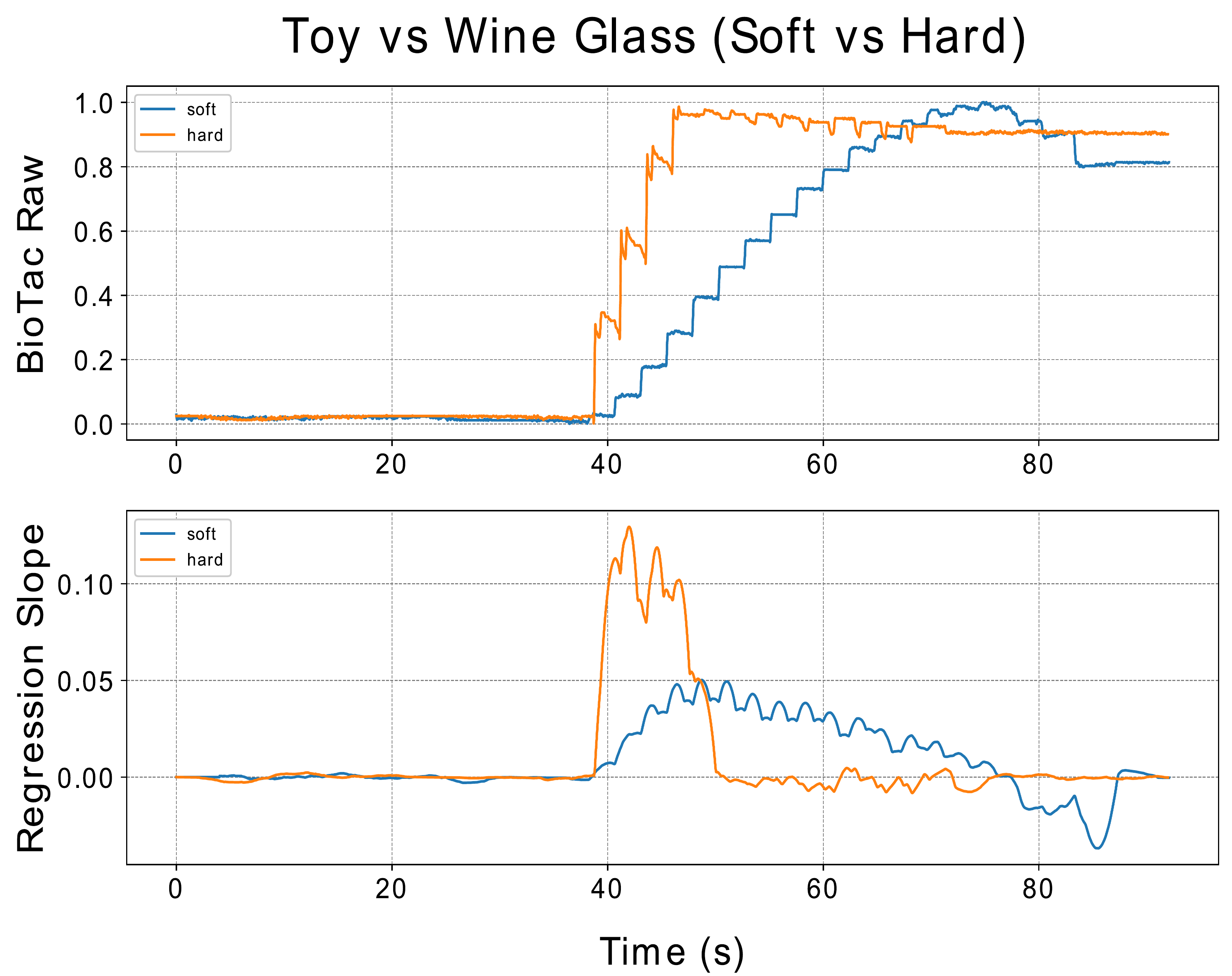}
    \caption{Variation of computed slope for soft and hard object. Notice the difference in the rate at which the slope changes.}
    \label{fig:slope_comparison}
\end{figure}

One of the crucial aspects of our proposed grasping pipeline is the ability to detect, and react to objects slipping between the fingers during grasp and move. This reactive nature of our controller allows for precise force applications on zero-shot objects, i.e., objects of unknown masses, sizes or shapes. 

For slippage detection, we utilize data from the BioTac sensor. Fig.~\ref{fig:grasp_slip_combo} show two sets of plots of sensors readings captured during grasping a wine glass, without and during slip respectively. From top to bottom, the graphs represent the BioTac, FSR readings and the position of the UR-10 along the z-axis. For visual clarity, we plot data for only the first finger. The difference in readings during slip versus without slip is quite evident, with several micro-vibrations in the BioTac data while the object slowly slips off the hand. This is due to the frictional properties of the BioTac skin, as well as the weight of the object. These vibrations are absent when the object does not slip, and the readings maintain a mostly stable baseline.
Our slip detection algorithm works by measuring and tracking the change in gradient of the sensor readings over a non-overlapping time-window of $\Delta t = 100ms$.

We use linear regression~\cite{linear_regression} to obtain a the instantaneous slope over the time-window, and perform the comparison at consecutive intervals, as shown in Fig.~\ref{fig:slope_comparison}. Consequently, by measuring the relative change in gradient, we are able to judge how fast the object is slipping, and provide larger or smaller control commands as necessary.  

\section{Experimental Results}
\label{sec:experiments}

\subsection{Experiment Setup}

The setup that we use to implement our pipeline includes multiple different robotic and sensing hardware, primarily the UR-10 manipulator and the Shadow Dexterous Hand, equipped with SynTouch BioTac tactile sensors~\cite{biotac}.
Our approach utilizes a switching controller architecture, where we deploy different strategies for controlling the UR-10 arm and the different joints of the Shadow Hand \cite{tuffield2003shadow} with feedback between controllers. The underlying control inputs come from various tactile sensors and the joint angles of the Shadow Hand.

We also introduce the \texttt{shadowlibs}\footnote{\href{https://github.com/kanishkaganguly/shadowlibs}{\faGithub \quad\texttt{kanishkaganguly/shadowlibs}}} library, a software toolkit that contains several utility functions for controlling the Shadow Hand.

\subsection{Results}
\label{sec:results}
\begin{figure}[t!]
    \centering
    \begin{subfigure}[b]{1.0\linewidth}
        \includegraphics[width=1.0\linewidth]{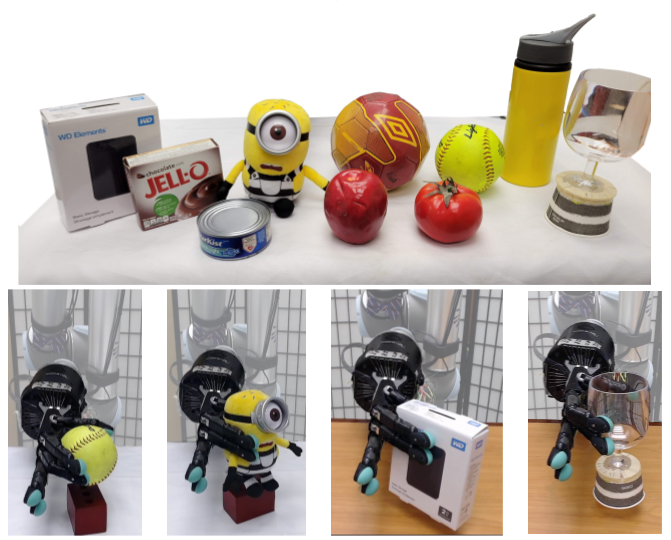}
        \caption{Top: Dataset of objects used in experiments.\\ Bottom (L-R): Grasping of Softball, Soft Toy, Box and Wine Glass.}
        \label{fig:experiments}
    \end{subfigure}
    
    \begin{subfigure}[b]{1.0\linewidth}
        \includegraphics[width=1.0\linewidth]{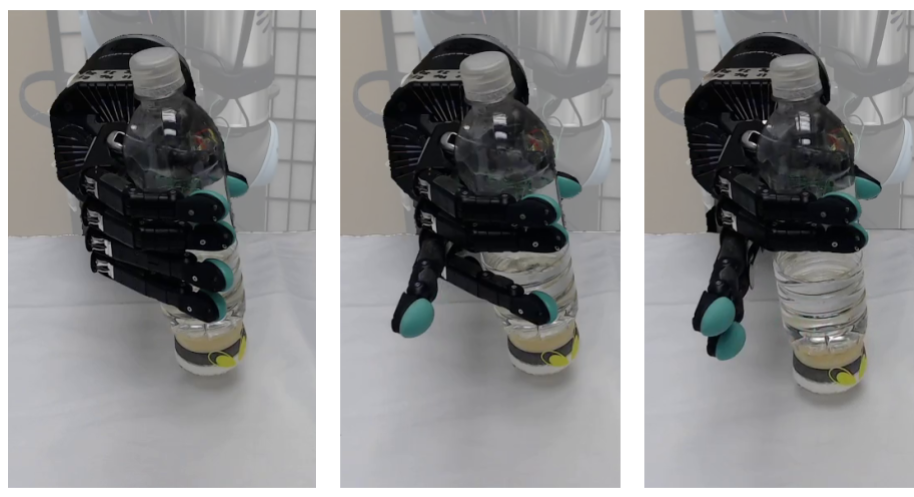}
        \caption{Demonstration of the stable grasp as the number of fingers contacting the object are reduced.}
        \label{fig:grasp_multifinger}
    \end{subfigure}
    
    \caption{Results from Experiments}
\end{figure}

We demonstrate our algorithm on a set of objects with varied shapes and sizes. The upper row of Fig.~\ref{fig:experiments} shows the dataset used, and the lower row of Fig. \ref{fig:experiments} shows successful grasps of four classes of objects, namely spherical, non-rigid, cuboidal, and transparent cylindrical respectively. We are able to grasp a wine glass, a soft toy, a ball and a box without any human intervention and without prior knowledge of their shapes, sizes or weights. The criteria for successful grasp were based on the ability to not only grasp the object entirely, but also to lift it and hold it in suspension for 10 seconds. Table~\ref{tbl:results} summarizes our results for various objects in our dataset. The accompanying video submission shows a detailed view of the grasping process, with cases for slip and without, as well as results for the other objects.

\begin{table}
\centering
\caption{Success rate over different object classes.}
\vspace{-5px}
\begin{tabular}{cc}
\toprule
\textbf{Objects} & \textbf{Success (\%)}     \\ \hline
Bottle                        & 85                        \\ 
Transparent Wine Glass                    & 80                        \\ 
Tuna Can                      & 70                        \\ 
Football                        & 60                        \\ 
Softball                        & 65                        \\ 
Jello Box                       & 70                        \\ 
Electronics Box                 & 85                        \\ 
Apple                        & 75                        \\ 
Tomato                       & 70                        \\
Soft Toy                  & 75                        \\ \bottomrule
\end{tabular}

\label{tbl:results}
\end{table} 


\begin{table}[!ht]
\setlength{\tabcolsep}{3pt}
\centering
\caption{Comparison with other state-of-the-art approaches.}
\begin{tabular}{ccccc}
\toprule
\textbf{Paper}                             & \textbf{Method} & \textbf{Num. Objects}   & \textbf{Success (\%)}  \\\hline
Li, \etal\cite{li2016dexterous}            & Simulation       & 3                       & 53.3               \\
Liu, \etal \cite{liu2020diffgrasping}       & Simulation       & 50                      & 66.0                \\
Saxena, \etal\cite{saxena2011dishwasher}   & Vision          & 9                       & 87.8                                      \\
Wu, \etal\cite{wu2019mat}                  & Reinf. Learning              & 10                      & 98.0              \\
Pablo, \etal\cite{pablo2019graspstability}                  & GCN              & 51                      & 76.6              \\
\hline
\textbf{Ours}                                        & Tactile based Control         & 10                      & 73.5              \\\bottomrule
\end{tabular}%
\label{tbl:compare}
\end{table}

\section{Analysis}
\label{sec:analysis}
As can be seen from Fig.~\ref{fig:slope_comparison}, there is a clear distinction between the BioTac's response to soft versus hard objects. The resulting slopes are also clearly distinguishable in their rate of change. Intuitively, softer objects slip slowly as compared to harder objects. This opens up interesting future avenues for material-adaptive grasping using our approach. 

We also demonstrate in Fig.~\ref{fig:grasp_multifinger}, that our controller is able to adjust to changes to the number of fingers in contact with the object on-the-fly. In particular, the object remains in a stable grasp even after two fingers are removed from contact showing the adaptive and robust nature of our approach. Such an approach has built-in recovery from possible failure of finger joints.

Lastly, as can be seen from the results, our proposed method, while simple, is quite adept at zero-shot object grasping. Compared to other methods, as shown in Table~\ref{tbl:compare}, such as those that use vision or learning, we are able to achieve comparable accuracy. Since our approach only relies on tactile data, we can also robustly grasp transparent objects with relative ease.

\section{Conclusions}
\label{sec:conclusions}
In this work, we develop a simple closed-loop formulation to grasp and manipulate a zero-shot object (object without a prior on shape, size, material or weight) with only tactile feedback. Our approach is based on the concept that we need to compensate for object slip to grasp correctly. We present a novel tactile-only closed-loop feedback controller to compensate for object slip. We experimentally validate our approach in multiple real-world experiments with objects of varied shapes, sizes, textures and weights using a combination of the Shadow Dexterous Hand equipped  with BioTac SP tactile sensors. Our approach achieves a success rate of 73.5\%. As a parting thought, our approach can be augmented by a zero-shot segmentation method \cite{NudgeSeg} to push the boundaries of learning new objects through interaction.

\newpage
\typeout{}
\bibliography{references}
\bibliographystyle{IEEEtran}

\end{document}